\newcommand{\CLASSINPUTtoptextmargin}{1.905cm}
\newcommand{\CLASSINPUTbottomtextmargin}{4.3cm}

\documentclass[conference, a4paper,10.9pt]{IEEEtran}

\usepackage{blindtext, graphicx}
\graphicspath{ {Images/} }

\usepackage[english]{babel}
\usepackage{comment}
\usepackage{array}
\usepackage[utf8]{inputenc} 
\usepackage{csquotes}

\usepackage{textcomp}
\usepackage{float} 
\def\BibTeX{{\rm B\kern-.05em{\sc i\kern-.025em b}\kern-.08em
    T\kern-.1667em\lower.7ex\hbox{E}\kern-.125emX}}
    
\usepackage[sorting=none]{biblatex}
\usepackage{ragged2e}

\ifCLASSINFOpdf
  
\else
  
\fi

\usepackage{balance}
\usepackage[T1]{fontenc}
\usepackage{mathptmx}
\usepackage{subcaption} 
\renewcommand{\figurename}{Fig.}
\addto\captionsenglish{\renewcommand{\figurename}{Fig.}}
\begin{document}
\title{\Huge{Benchmarking Conventional Vision Models on Neuromorphic Fall Detection and Action Recognition Dataset}}
\author{
\IEEEauthorblockN{\small{Karthik Sivarama Krishnan}}
\IEEEauthorblockA{
\small{Email: ks7585@rit.edu}}
\and
\IEEEauthorblockN{\small{Koushik Sivarama Krishnan}}
\IEEEauthorblockA{
\small{Email: koushik.nov01@gmail.com}}}

\maketitle

\begin{abstract}
    Neuromorphic vision-based sensors are gaining popularity in recent years with their ability to capture Spatio-temporal events with low power sensing. These sensors record events or spikes over traditional cameras which helps in preserving the privacy of the subject being recorded.
    These events are captured as per-pixel brightness changes and the output data stream is encoded with time, location, and pixel intensity change information. This paper proposes and benchmarks the performance of fine-tuned conventional vision models on neuromorphic human action recognition and fall detection datasets. The Spatio-temporal event streams from the Dynamic Vision Sensing cameras are encoded into a standard sequence image frames. These video frames are used for benchmarking conventional deep learning-based architectures. In this proposed approach, we fine-tuned the state-of-the-art vision models for this Dynamic Vision Sensing (DVS) application and named these models as  DVS-R2+1D, DVS-CSN, DVS-C2D, DVS-SlowFast, DVS-X3D, and DVS-MViT. Upon comparing the performance of these models, we see the current state-of-the-art MViT based architecture DVS-MViT outperforms all the other models with an accuracy of 0.958 and an F-1 score of 0.958. The second best is the DVS-C2D with an accuracy of 0.916 and an F-1 score of 0.916. Third and Fourth are DVS-R2+1D and DVS-SlowFast with an accuracy of 0.875 and 0.833 and F-1 score of 0.875 and 0.861 respectively. DVS-CSN and DVS-X3D were the least performing models with an accuracy of 0.708 and 0.625 and an F1 score of 0.722 and 0.625 respectively.\\
\end{abstract}

\IEEEpeerreviewmaketitle

\section{\normalsize{Introduction}}
In recent years, there has been a significant increase in human monitoring systems. From wearable devices to surveillance cameras to smartphones etc, every device is collecting data about the user and stores it in online database systems. With all these advancements in the field of Artificial intelligence, there is a decline in privacy and information protection of every human being.

In the public healthcare domain, a Fall is considered one of the major emergency events in the world. According to the World Health Organization \cite{WHO}, an estimated 684,000 individuals die every year from fall events. According to the authors of the paper \cite{el2013fall}, it is estimated that by the year 2050, there will be more than one in every group of five people will be more than sixty-five years old. This is a concerning number when considering the growing counts of elderly people and the need for accurate and safe fall detection and prevention systems. With this increase in the rising elderly population, the health care for the elderly is increasing in massive numbers and would become one of the major world economic sectors \cite{s21030947}. This has attracted the attention of many researchers and research organizations over the last decades. 

Detecting human activity and fall has become a substantial domain of interest for researchers in the healthcare domain. There are multiple ways of recognizing human activities \cite{hussain2019different}. This includes, 
\begin{itemize}
    \item \textit{Radio Frequency based approaches} -  where radio signals are used to recognize human activities
    \item \textit{Sensor based approach} - where sensors like accelerometer, gyroscope, biosensors, pressure sensor, proximity sensor, etc. are used to recognize human activity through wearable/devices.
    \item \textit{Vision based approach} - where camera based sensors are used to detect activity patterns. 
\end{itemize}
   With the advancements in Deep learning based networks, there have been a huge increase in body worn sensors based approaches in detecting human activities \cite{wang2019deep}.On the other hand, vision based systems have become a prominent and robust solution in detecting a fall and other activity patterns.  \\   

The main concern with the vision-based systems is that, although these systems have great performance in recognizing human activities, there is a huge decline in the privacy of the subject that is monitored using these systems. In recent years, there has been a huge increase in the use of event-based cameras. These cameras differ from the traditional camera frames by asynchronously measuring the brightness changes at every pixel instead of capturing images at a standard rate\cite{s21030947}. 

Event based sensors can perform with extremely low latency, requiring very low computational power. The event based DVS camera used here is called the DAVIS346redColor neuromorphic vision camera. The 3 Dimensional video frames from this camera is converted into 2 Dimensional video frames for using the conventional video classification models for comparing and benchmarking the performance. Conventional video classification models shows extraordinary performance while classifying 2 Dimensional video frames. But the Dataset extraced from the neuromorphic vision camera consists of 3 Dimensional characteristics and are not suitable to be benchmarked with the current state of the art approaches. 

In this paper, we converted the neuromorphic vision camera frames into the classic 2 Dimensional frames and  we fine-tuned and retrained the standard video classification architectures that includes  R(2+1)D\cite{tran2018closer},  X3D\cite{feichtenhofer2020x3d}, MViT\cite{fan2021multiscale}, C2D\cite{zheltonozhskii2021contrast}, CSN\cite{tran2019video} and SlowFast ResNet50\cite{feichtenhofer2019slowfast}   to classify the actions performed in the video frames of neuromorphic vision datasets and also detect a fall. We then combined the datasets to include both Action Recognition and Fall Detection to make the model more robust in classifying both the tasks.

\section{\normalsize{Related Work}}

\normalsize{\emph{\subsection{Fall Detection and Action Recognition}}}
  Recognizing human actions have become a significant part of the research in recent times. The authors of the paper \cite{8250154} proposed a Support Vector Machine (SVM) machine learning based approach in recognizing human actions using a wearable sensor. The authors Jindong et al., of the survey paper \cite{wang2019deep} summarizes different approaches for sensor based activity recognition using Deep learning based approaches. The authors Javier et al., \cite{10.1007/978-3-642-22362-4_19} proposed a dynamic sliding window based approach where the window size and the time shift are not fixed at every step and are adjusted dynamically.  
\normalsize{\emph{\subsection{Neuromorphic Dataset}}}   
Event based cameras are gaining attention in the research community. The authors of the review paper \cite{gallego2019event} explains various optical flow based approaches to track the events over time. The authors of the paper \cite{8753848} propose a deep learning based appraoch for tracking the features over time and estimating the optical flow. Authors Salah et al.,\cite{al2021making} proposes the use of support vector machine (SVM) and K-Nearest Neighbours (KNN) based machine learning approach for detecting human action on Neuromorphic vision datasets. 

Authors He et al.,\cite{he2020comparing} proposed an approach comparing the Recurrent Neural Network (RNN)s performance over Spiking Neural Network (SNN) on neuromorphic vision dataset. 
\normalsize{\emph{\subsection{Video Classification}}}
  With the advancements in deep learning \cite{krishnan2021vision}, there has been a significant performance by the deep learning-based architectures in training and classifying actions that are similar to each other.
 \subsubsection{R(2+1)D}  The authors Tran et.al \cite{tran2018closer} proposed a 3D Convolutional Neural Network-based architecture that refines the spatial and temporal components to yield a significant performance in accuracy gains over the standard 2D Convolutional Neural Network for recognizing actions on video frames. With this approach the authors were able to achieve two advantages, By separating the spatial and temporal components the optimization of the model was easier and maintains the number of parameters by increasing the non-linearity.
 \subsubsection{CSN} The authors Tran et al, \cite{tran2019video} proposed an updated 3D Channel Separated Convolutional Neural Network architecture, that helped in substantial improvement in performance and thereby reducing the cost of computation. The performance was achieved by decomposing the convolution channels to either 1×1×1 conventional convolutions or k×k×k depthwise convolution which would help in separating channel interactions from spatiotemporal interaction. Similar channel separation networks have shown improved performance and lowered computation time \cite{krishnan2021swiftsrgan}. 
 \subsubsection{C2D}  The authors Evgenii et al, \cite{zheltonozhskii2021contrast} proposed a semi-supervised learning-based approach to improve the performance of learning with noisy labels (LNL) methods. With this approach, the authors were able to achieve robust performance on a severe noisy dataset. 
 \subsubsection{SlowFast} The authors Christoph et.al, \cite{feichtenhofer2019slowfast} proposed an SlowFast network architecture operating at two different frame rates. The path with a lower frame rate extracts the spatial features from the sequence and the path with a higher frame rate extracts the temporal features from the sequence. This approach helps in achieving higher performance in action recognition in videos. 
 \subsubsection{X3D} The author Christoph \cite{feichtenhofer2020x3d} proposed an efficient network that expands a single axis at every step. Thereby maintaining the accuracy performance and maintaining the model lightweight. 
\subsubsection{MViT} The authors Haoqi et.al, \cite{fan2021multiscale} proposed a vision transformer-based approach for video recognition. This approach uses multiscale features where the initial input resolution is smaller and expands the capacity of the channel which helps in reducing the spatial features. This approach helps in learning low dimensional features in the initial layers and complex high dimensional features in the deeper layers. By this approach, the authors were able to achieve higher accuracy on video recognition tasks.

\section{\normalsize{Methodology}}
\begin{figure*}[tp]
\vspace*{25pt}
\centering

\subfloat[Gamma Contrast]{\includegraphics[width=0.20\textwidth]{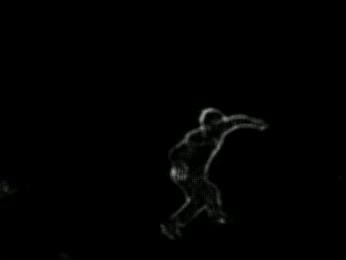}}\ 
\subfloat[Frequency Encoding]{\includegraphics[width=0.20\textwidth]{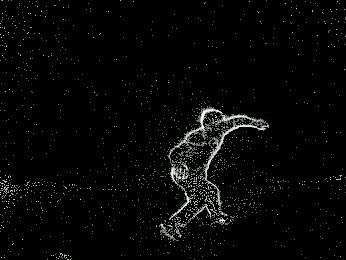}}\ 
\subfloat[Gaussian Blur]{\includegraphics[width=0.20\textwidth]{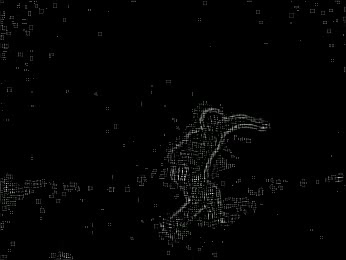}}\ 
\subfloat[Kernel Edges]{\includegraphics[width=0.20\textwidth]{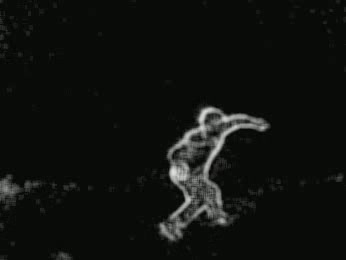}}\ 
\subfloat[CLAHE]{\includegraphics[width=0.20\textwidth]{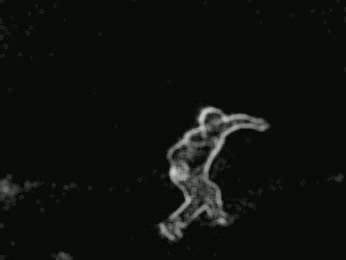}}\ 
\subfloat[Histogram Equalization]{\includegraphics[width=0.20\textwidth]{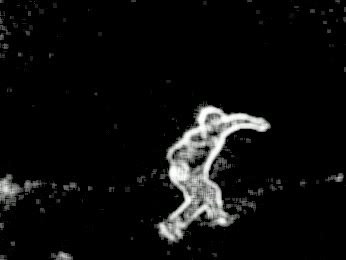}}\ 
\subfloat[Morphological Transformation]{\includegraphics[width=0.20\textwidth]{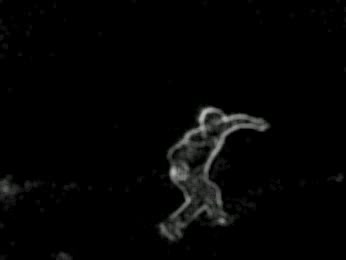}}\ 
\subfloat[SAE]{\includegraphics[width=0.20\textwidth]{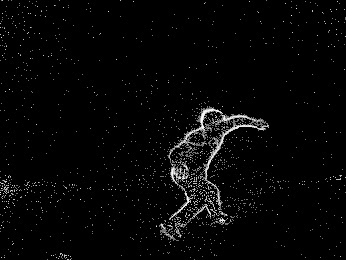}}\ 
\caption{Preprocessing and Data Augmentation sample frame on a Fall action Sequence.}
\label{fig:preProcess}

\end{figure*}
\normalsize{\emph{\subsection{Dataset}}}
The standard benchmarking neuromophic vision dataset \cite{miao2019neuromorphic} is considered here for the evaluation of the performance of MViT architecture. The dataset is recorded using the DAVIS346redColor neuromorphic vision camera. The Neuromorphic camera collects the spatio-temporal event stream consisting of 4 data-points for every event. These data-points consists of Timestamps, X-Coordinates, Y-Coordinates and Polarity information. 

\normalsize{\emph{\subsection{Dataset Preprocessing}}}
The asynchronous event streams are converted to image frames using the encoding techniques. Encoding techniques help create meaningful image sequences that can be processed and interpreted using standard deep learning techniques. 

\subsubsection{Frequency Encoding}
The frequency encoding technique is used to cancel out the noise from the sensor data by using the polarity information in the data. Given a pixel location (x, y), the frequent occurrences of positive and negative events are accumulated over a given interval and the value of the pixel is computed using the range normalization technique\cite{chen2018pseudolabels}. This helps in improving the edges of the object present in the image. A sample image for this encoding technique is shown at Figure \ref{fig:preProcess} (b)

\subsubsection{Surface of Active Events (SAE)}
Now utilizing the timestamp information from the data, the Surface of Active Events (SAE) approach helps in changing the pixel value at every (x, y) location according to timestamp t. This generates a grayscale image corresponding to the timestamp of the event at every pixel \cite{2017}. Numerical mapping is then applied to generate an 8-bit single-channel image. A sample image for this encoding technique is shown at Figure \ref{fig:preProcess} (h)
\normalsize{\emph{\subsection{Feature extraction}}}
The fall detection dataset consists of 4 classes which include, falling down, picking up, sit-down and tying shoes. The action recognition dataset consists of 10 classes that include arm-crossing, getting up, jumping, kicking, picking up, sit-down, throwing, turning around, walking, and waving. Since both datasets consist of 2 overlapping classes, these two were considered once and the total classes for our combined dataset come to be 12.  The actions on these datasets are performed on a subset in the sequences for a duration of about 5 seconds. These features representing the action class are extracted from the dataset and the video frames are prepared for the training.  This helps the model to understand and classify the fall and other actions with greater accuracy. We randomly chose a sequence of 16 frames from a total of 25(FPS) * 3(duration) frames.

\normalsize{\emph{\subsection{Data Augmentation}}}
The combined dataset consists of 30 recordings for every class with a 5 seconds average length of actual action. since the amount of data is small, the combined dataset has to be augmented in order to prepare more synthetic data from the actual dataset. This also helps in preventing the overfitting of the model. The combined dataset is augmented by adjusting the gamma contrast by scaling pixels value Figure \ref{fig:preProcess} (a), equalizing the histogram within image patches by using the CLAHE (Contrast Limited Adaptive Histogram Equalization) technique Figure \ref{fig:preProcess} (e), applying edge detection to detect the region around the object, and applying a gaussian blur Figure \ref{fig:preProcess} (c) to reduce the excessive noise in the combined dataset. These augmented sequences are done across all the classes in the training and validation datasets thereby increasing the fall detection dataset. Figure \ref{fig:preProcess} shows various samples outputs from the feature extraction and data augmentation samples.
     \begin{table}
     \centering
    \caption{Dataset Split} 
    \label{table:data}
    \renewcommand*\arraystretch{1.6}
    \normalsize
   \begin{tabular}{|>{\centering\arraybackslash}m{1in}|>{\centering\arraybackslash}m{2in}|}
    
    \hline
    Video Sequence & Combined Fall Detection + Action Recognition  \\
    \hline 
    Train    & 4328\\
    Validation  & 1072\\
    Test  & 24\\
    Total   & 5424\\
    \hline
    \end{tabular}
    \end{table}
\\
\section{\normalsize{Experiments and Analysis}}
\normalsize{\emph{\subsection{Experimental Setup}}}
    We tested out R(2+1)D\cite{tran2018closer}, SlowFast\cite{feichtenhofer2019slowfast}, X3D\cite{feichtenhofer2020x3d}, MViT\cite{fan2021multiscale}, C2D\cite{zheltonozhskii2021contrast}, CSN\cite{tran2019video} and SlowFast ResNet50\cite{feichtenhofer2019slowfast} pre-trained models and fine-tuned it on the combined dataset. In transfer-learning, all other layers except the last one is frozen and the last layer's output neuron is updated based on the number of classes on the datasets. We combined the classes of both the action recognition and fall detection datasets. So the total classes (output layer of the model) is 12. The table \ref{table:data} describes the distribution of training , validation and test sets.\\
    
    We chose these six models in specific as they have shown promising results when fine-tuning on various video datasets. These models in general have achieved state-of-the-art results on video action recognition tasks. Hence we finetuned and benchmarked them on the neuromophic vision dataset \cite{miao2019neuromorphic}.\\
    
    We experimented with various optimizers, learning rate, and activation functions between the last linear layers using Optuna framework\cite{akiba2019optuna} to find the best combination. We conducted 50 experiments with different combinations of the optimizers, learning rate, and activation functions for each model. The following are the best optimizer and activation functions for each model the yielded the best performance.\\
    
    The best X3D\cite{feichtenhofer2020x3d} model was trained with AdamW and ReLU activation between the linear classifier layers, the best MViT\cite{fan2021multiscale} model was with Rectified Adam(RAdam)\cite{liu2021variance} that rectifies the variance of adaptive learning rate and relu activations between the linear classifier layers, the best C2D\cite{zheltonozhskii2021contrast} model was with RAdam\cite{liu2021variance} optimizer and GELU activation between the last layers, the best CSN\cite{tran2019video} model was with RAdam\cite{liu2021variance} optimizer and ReLU activation between the last linear layers and, the best SlowFast ResNet50\cite{feichtenhofer2019slowfast} model was trained with Adam and GELU activation between the last linear layers.\\
    
\begin{figure}[!ht]
    \vspace*{20pt}
    \includegraphics[width = 8.2cm, height = 5cm]{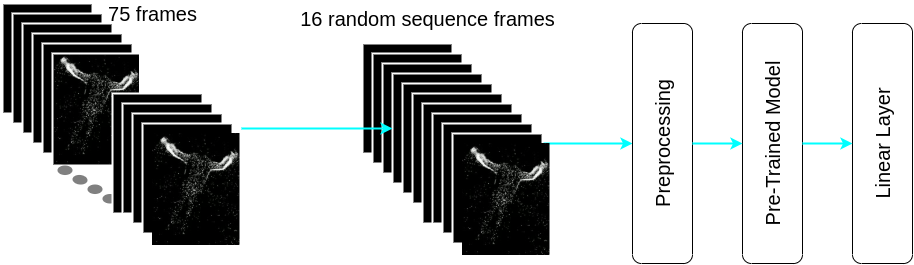}
    \caption{\footnotesize{Model Architecture}}
    \label{fig:subim6}
    \end{figure}

    All four models are trained using PyTorch on RTX 2070 Super GPU with gradient accumulation and mixed precision. Early Stopping and Stochastic Weight Averaging are used for preventing the network from over-fitting. Figure \ref{fig:subim6} shows the transfer learning architecture used to evaluate multiple model architectures. \\
    
    For the dataset, we used a sliding window-based approach considering 3(seconds) of data with 25 frames per second took a random 16 sequence frame from it. We also used various augmentation techniques like Gamma Contrast, Histogram Equalization, CLAHE and, Gaussian Blur.

\normalsize{\emph{\subsection{Evaluation Criteria}}}
    To evaluate the performance of our models, we used Accuracy, Precision, Recall and F1-score. All these metrics put together, can accurately depict the model's real-world performance.
    
    \begin{equation}
    Accuracy  = \frac{TP+TN}{TP+FP+FN+TN}
    \end{equation}
    \begin{equation}
        Precision = \frac{TP}{TP+FP}
    \end{equation}
    \begin{equation}
        Recall = \frac{TP}{TP+FN}
    \end{equation}
    \begin{equation}
        F1-score = 2 * \frac{Recall * Precision}{Recall + Precision}
    \end{equation}

\section{\normalsize{Results}}
         \begin{table} [ht]
    \caption{Performance} 
    \label{table:performance}
    \renewcommand*\arraystretch{2}
    \normalsize
  \noindent \begin{tabular}{|p{13.5mm}|p{12.5mm}|p{12.5mm}|p{8.5mm}|p{8.5mm}|p{8.5mm}|}
    
    \hline
    \textbf{Data} & \textbf{Model} & \textbf{Precision} & \textbf{Recall} & \textbf{F1} & \textbf{ACC}\\
    \hline

    Combined  & DVS-C2D    & 0.916 & 0.916 & 0.916 & 0.916\\
   (Fall  & DVS-SlowFast       & 0.916 & 0.833 & 0.861 & 0.833\\
      +   & DVS-CSN      & 0.750 & 0.708 & 0.722 & 0.708 \\
   Action) & DVS-X3D  & 0.645 & 0.625 & 0.625 & 0.625\\
   Dataset & DVS-R2+1D & 0.875 & 0.875 & 0.875 & 0.875 \\
    & \textbf{DVS-MViT}   & \textbf{0.958} & \textbf{0.958} & \textbf{0.958} & \textbf{0.958}\\
    \hline
    \end{tabular}
    \end{table}

    The above table \ref{table:performance} shows the performance of the models with the best performing optimizer and activation function combination. Since these models are fine-tuned, trained, and evaluated for Dynamic Vision Sensing (DVS) datasets, we named the new models as DVS-C2D, DVS-SlowFast, DVS-CSN, DVS-X3D, DVS-R2+1D, and DVS-MViT. Clearly, from the above table, DVS-CSN and DVS-X3D were the least performing models with an accuracy of 0.708 and 0.625 and an F1 score of  0.722 and 0.625 respectively. The DVS-C2D and DVS-R2+1D rank in the second and third places with an accuracy of 0.916 and 0.875 and F1 score of 0.916 and 0.875 respectively and DVS-SlowFast on the fouth spot with an accuracy of 0.833 and F1 score of 0.861. From the table above, we can infer that the DVS-MViT model was the best performing model with an accuracy of 0.958, precision of 0.958, recall of 0.958 and, f1-score of 0.958. Multivariate Vision Transformer-based architecture is the current state-of-the-art approach for video classification applications. Hence, we see a similar performance on our application too.

\section{\normalsize{Conclusion and Future Work}}
In this work, we demonstrated the process of using conventional video classification models for the Neuromorphic Vision dataset. We fine-tuned six state-of-art conventional video classification models and benchmarked the performance of these models into the combined dataset from action recognition and fall detection. Since this is the first of a kind approach, we compared the performance among other models to report the best-performing architecture.

    The DVS-MViT \cite{fan2021multiscale} model was the best performing on the combined fall detection and action recognition dataset. The DVS-MViT achieved an accuracy of 0.958 and an F1-score of 0.958. We achieved the new state-of-the-art in both action recognition and fall detection by fine-tuning the MViT model with RAdam\cite{liu2021variance}. MViT is the current state-of-the-art model on video classification tasks. In future work, we are planning to develop a new architecture that can work in real-time with low latency and low computational capabilities and benchmark the performance and computation time along with these architectures.

\ifCLASSOPTIONcaptionsoff
  \newpage
\fi

\balance
\printbibliography
\end{document}